\documentclass{article}
\usepackage{spconf,amsmath,amssymb,graphicx,hyperref}
\usepackage{subcaption}
\usepackage{enumitem}
\usepackage{booktabs}

\title{Subtractive Modulative Network with Learnable Periodic Activations}
\name{Tiou Wang$^{\ast}$, Zhuoqian Yang$^{\dagger}$, Markus Flierl$^{\ast}$, Mathieu Salzmann$^{\dagger}$, Sabine Süsstrunk$^{\dagger}$}
\address{
    $^{\ast}$School of Electrical Engineering and Computer Science, KTH Royal Institute of Technology, Sweden \\
    $^{\dagger}$School of Computer and Communication Sciences, EPFL, Switzerland \\
    $^{\ast}$\{tiou, mflierl\}@kth.se, $^{\dagger}$first.last@epfl.ch
}

\begin{document}
%
\maketitle
\begin{abstract}
We propose the Subtractive Modulative Network (SMN), a novel, parameter-efficient Implicit Neural Representation (INR) architecture inspired by classical subtractive synthesis. The SMN is designed as a principled signal processing pipeline, featuring a learnable periodic activation layer (Oscillator) that generates a multi-frequency basis, and a series of modulative mask modules (Filters) that actively generate high-order harmonics. We provide both theoretical analysis and empirical validation for our design. Our SMN achieves a PSNR of $40+$ dB on two image datasets, comparing favorably against state-of-the-art methods in terms of both reconstruction accuracy and parameter efficiency. Furthermore, consistent advantage is observed on the challenging 3D NeRF novel view synthesis task. Supplementary materials are available at \url{https://inrainbws.github.io/smn/}.
\end{abstract}
\section{Introduction}
\label{sec:intro}

Implicit Neural Representations (INRs) have emerged as a powerful method for continuous signal representation using coordinate-based networks~\cite{tancik2020fourier,sitzmann2020siren,tancik2021learned}. However, their practical efficacy is often hindered by a fundamental limitation known as spectral bias~\cite{pmlr-v97-rahaman19a}, where standard multilayer perceptron (MLP) architectures struggle to learn high-frequency content, leading to blurry reconstructions and slow convergence. While Fourier feature mappings can mitigate this, they often result in monolithic black box models where spectral components are entangled and inefficiently combined through simple additive synthesis.

To address these challenges of inefficiency and interpretability, we propose the \textbf{Subtractive Modulative Network (SMN)}, an INR architecture conceptually inspired by the principles of subtractive synthesis from classical signal processing. Subtractive synthesis is an audio signal processing technique where waveforms are shaped by filters to remove frequencies and create the desired. Instead of treating the network as a monolithic function approximator, we structure it as a multi-stage signal processing pipeline. This framework is composed of an Oscillator stage, realized as a learnable periodic activation layer that generates a multi-frequency basis, and a series of Filter stages, which are implemented with Modulative Mask modules.

Our contributions are: \textbf{(1)} We introduce a \textbf{Learnable Sine Layer} as an adaptive ``Oscillator''. We show that this compact, learnable activation function, where the addition of just a few parameters can yield a performance gain of $7 \sim 9$ dB on a high-fidelity baseline, providing a more efficient and effective frequency basis than standard fixed encodings for 2D signals. \textbf{(2)} We propose a series of Modulative Mask modules that employ \textbf{multiplicative interactions}. We provide both theoretical and empirical evidence that this mechanism is a fundamentally superior method for harmonic generation and spectral sculpting compared to simple addition.

\section{RELATED WORK}
\label{sec:related_work}

Implicit Neural Representations (INRs) model a continuous signal as a function $f_\theta: \mathbb{R}^d \rightarrow \mathbb{R}^c$ implemented by a coordinate-based MLP. A key challenge in designing INRs is overcoming the inherent spectral bias of standard MLPs, which struggle to learn high-frequency functions~\cite{pmlr-v97-rahaman19a}. This phenomenon has been more broadly analyzed in terms of the \emph{Frequency Principle}, which shows that neural networks tend to fit low-frequency components of a target function before higher ones~\cite{xu2019frequency}. Existing approaches to mitigate this bias can broadly be divided into two categories: input feature mapping and the use of periodic activation functions.\par

Two primary strategies enable neural networks to represent high-frequency signals. One approach maps low-dimensional input coordinates $\mathbf{x}$ to a higher-dimensional feature space using a fixed sinusoidal basis, such as the positional encoding $\gamma(\mathbf{x}) = [\sin(\mathbf{B}\mathbf{x}), \cos(\mathbf{B}\mathbf{x})]$ used in Fourier Feature Networks~\cite{tancik2020fourier} and NeRF~\cite{mildenhall2021nerf}. While effective, this method offers limited explicit control over the signal's spectral composition. An alternative paradigm, pioneered by SIREN~\cite{sitzmann2020siren}, employs sinusoidal activations throughout the network to create a strong inductive bias for signals with smooth derivatives. This concept was further advanced by models like WIRE~\cite{saragadam2023wire}, which uses wavelet-based activations to improve the spatial-spectral localization trade-off, highlighting the critical role of activation function design.



\subsection{Subtractive Synthesis and Modulation} 

Subtractive synthesis is a signal generation technique that sculpts a desired output by filtering a spectrally rich source signal, $x(t)$, in contrast to additive synthesis which sums individual frequency components. Originating in analog synthesizers, this method is formally expressed in the frequency domain as the element-wise product $Y(\omega) = X(\omega) \cdot H(\omega)$, where the filter $H(\omega)$ attenuates unwanted harmonics. 

Most architectures implicitly adhere to an additive synthesis~\cite{chen2019learning,dupont2021coin} paradigm. In this model, signal components, often derived from Fourier features or periodic activations, are incrementally superimposed layer by layer. Although effective, this additive approach can be inefficient, as the network may need to learn complex cancellations to eliminate undesired harmonics~\cite{damodaran2023improved}. A subtractive framework offers a promising alternative. By starting with a high-frequency embedding of an input coordinate, $\gamma(\mathbf{x})$, a subtractive INR can employ a learned masking function, $\mathcal{M}$, to selectively suppress spectral content, yielding the final representation $f_\theta(\mathbf{x}) = \mathcal{M}(\gamma(\mathbf{x}))$. This approach enables more direct control over spectral structure, potentially improving efficiency and mitigating the limitations of additive models. Recent proposals, including frequency-aware masking~\cite{dupont2021coin} and wavelet-based decomposition~\cite{skorokhodov2021adversarial}, can be interpreted as early attempts toward this paradigm.



\section{Method}
\label{sec:MD}

\begin{figure}[t]
    \centering
    \includegraphics[width=\linewidth]{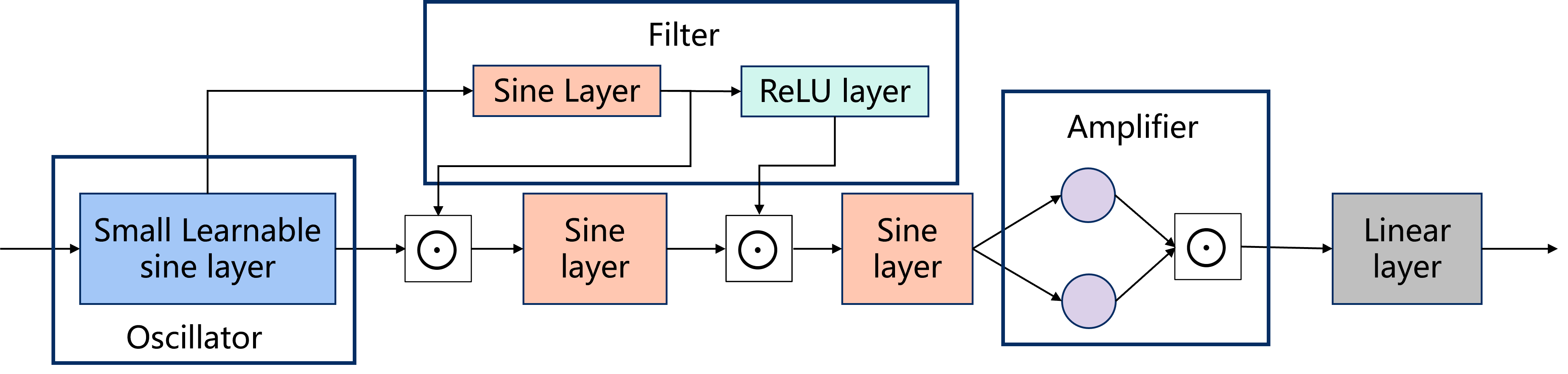} 
    \caption{The end-to-end architecture of our SMN, illustrating the Oscillator, the multi-stage Filter with its main and masking pathways, and the final Amplifier (Self-Mask) stage.}
    \label{fig:network_architecture}
    \vspace{-1em}
\end{figure}

We introduce the Subtractive Modulative Network (SMN), an INR architecture conceptually inspired by the subtractive synthesis in audio signal  processing~\cite{Barkan2019}. Unlike standard monolithic MLPs, our framework decomposes the signal synthesis process into a structured, multi-stage pipeline comprising an ``Oscillator'' for frequency generation and a ``Filter'' for spectral sculpting which are both stages of subtractive synthesis. The complete architecture is illustrated in Figure~\ref{fig:network_architecture}.

\subsection{The Oscillator: A Learnable Sine Layer}
\label{subsec:oscillator}

To overcome the spectral bias of standard networks, the Oscillator stage is designed to generate a rich, multi-frequency basis at the first layer. It is implemented as a linear layer followed by a custom \textit{learnable periodic activation function}, $\Phi(\cdot)$. For an input coordinate $\mathbf{x} \in \mathbb{R}^d$, the process is twofold:
\begin{align}
    \mathbf{v} &= \mathbf{W}_0 \mathbf{x} + \mathbf{b}_0, \\
    \mathbf{z}_{osc} &= \Phi(\mathbf{v}; \mathbf{a}) = \sum_{i=1}^{K} a_i \sin(\omega_i \mathbf{v}),
    \label{eq:learnable_activation_final}
\end{align}
where $\mathbf{W}_0, \mathbf{b}_0$ are standard layer parameters, $\{\omega_i\}_{i=1}^K$ is a set of fixed, multi-resolution frequencies (e.g., $\{8, 40,120\}$), and $\mathbf{a} = \{a_i\}_{i=1}^K$ is a vector of learnable scalar amplitudes. This design allows the network to adaptively learn the optimal mixture of basis frequencies for a given signal by adjusting the coefficients $a_i$, providing a more efficient and effective spectral basis than fixed encodings.

\subsection{The Filter: A Multi-Stage Modulative Mask}
\label{subsec:filter}

The Filter stage is the core of our framework, realized through a novel multi-stage modulation mechanism. Its design is based on a key theoretical insight: multiplicative interactions are fundamentally superior for generating new harmonics, a crucial capability for representing complex signals. We discuss this further in the supplementary materials. The composition of sine activations, as in $\sin(\sin(\omega z))$, implicitly generates an infinite series of higher-order harmonics ($3\omega, 5\omega, \dots$), providing a powerful mathematical foundation for our model's expressivity.

Our implementation involves a main signal pathway and a parallel masking pathway. Let the Oscillator's output be $\mathbf{z}^{(1)}$.
\begin{enumerate}[leftmargin=*]
    \item \textbf{Initial Additive Modulation:} A masking signal $\mathbf{m}^{(1)} = \sin(\mathbf{W}_{mask}^{(1)}\mathbf{z}^{(1)} + \mathbf{b}_{mask}^{(1)})$ is generated and additively combined with the main path signal: $\hat{\mathbf{z}}^{(1)} = \mathbf{z}^{(1)} + \mathbf{m}^{(1)}$. The main pathway then produces the next-stage feature $\mathbf{z}^{(2)} = \sin(\mathbf{W}_{main}^{(2)}\hat{\mathbf{z}}^{(1)} + \mathbf{b}_{main}^{(2)})$.
    
    \item \textbf{Predictive Multiplicative Masking:} The masking pathway evolves by generating a multiplicative mask $\mathbf{M}^{(2)}$ from the first modulation signal. This mask is then applied to the main pathway's signal in a form of one-layer-ahead predictive filtering:
    \begin{align}
        \mathbf{m}^{(2)} &= \sin(\mathbf{W}_{mod}^{(2)}\mathbf{m}^{(1)} + \mathbf{b}_{mod}^{(2)}), \\
        \mathbf{z}^{(3)} &= \mathbf{z}^{(2)} \odot \mathbf{m}^{(2)}.
    \end{align}
    This multiplicative step is where the primary spectral sculpting occurs. $\odot$ is element-wise multiplication. The first $\odot$ can be replaced by element-wise add. If the small learnable layer is treated as the encoding part, here $\odot$ and element-wise add are the same in mathematic sense.

    \item \textbf{Self-Mask Amplifier:} After several such layers, a final Self-Mask, implemented as an element-wise squaring operation ($\mathbf{z}_{final} = (\mathbf{z}^{(L)})^2$), serves as a parameter-free ``Amplifier'' that enhances non-linearity and generates second-order harmonics~\cite{proakis2001digital}.
\end{enumerate}

\subsection{Overall Architecture and Training}
\label{subsec:training}

The complete SMN is a feed-forward network with 4 hidden layers (including mask layers), each with 256 hidden units. A final linear layer maps the features to the output dimension. The network's parameters $\theta$ are optimized end-to-end by minimizing the Mean Squared Error (MSE) loss between the network output $\mathcal{F}_\theta(\mathbf{x}_i)$ and ground truth values $\mathbf{y}_i$:
\begin{equation}
    \mathcal{L}(\theta) = \frac{1}{N} \sum_{i=1}^{N} \left\| \mathcal{F}_\theta(\mathbf{x}_i) - \mathbf{y}_i \right\|_2^2.
\end{equation}
We employ the Adam optimizer~\cite{kingma2015adam} with an initial learning rate of $2 \times 10^{-2}$ and a \texttt{ReduceLROnPlateau} scheduler that halves the learning rate if the loss plateaus for 100 iterations. Models are trained for a maximum of 5,000 iterations.

\section{EXPERIMENTS}
\label{sec:exp}

We evaluate the performance on 2D high-fidelity image representation against state-of-the-art baselines. We then assess its generalization capabilities on the more challenging task of 3D novel view synthesis. Finally, a series of ablation studies are presented to analyze the contribution of our key architectural components: oscillator and filter design.

\subsection{Experimental Setup}
\label{subsec:setup}

For 2D image representation, we use two standard benchmarks: the 24 high-resolution images from the \textbf{Kodak} dataset whose resolution is $768 \times 512$ or $512 \times 768$, and the first 24 $510 \times 339$ images from the \textbf{DIV2K} validation set (LR mild track)~\cite{agustsson2017ntire}, which represent a different data distribution. For 3D novel view synthesis, we use the eight scenes of the synthetic \textbf{NeRF} dataset at a 400$\times$400 resolution.

We compare our SMN against prominent INR architectures, including Gauss~\cite{ramasinghe2021beyond} with $s_0=30$, SIREN~\cite{sitzmann2020siren} with $\omega_0=40$, WIRE~\cite{saragadam2023wire} with $\omega_0=20$ and $s_0=10$, following the metrics in WIRE~\cite{saragadam2023wire}, and Robust Filters (RINR)~\cite{ma2025robustifying}. For the NeRF task, all models are integrated with a standard Positional Encoding (PE) layer~\cite{mildenhall2021nerf} for fair comparison. Performance is primarily evaluated using Peak Signal-to-Noise Ratio (PSNR) and model parameter count. While architectural differences prevent an exact parameter match, we kept the counts as close as possible.

\subsection{Main Results on 2D Image Representation}
\label{subsec:results_2d}

We first benchmark all models on the task of single-image fitting. Table~\ref{tab:2d_results} presents the average PSNR and parameter counts for both the Kodak and DIV2K datasets.

\begin{table}[t]
    \centering
    \caption{Quantitative comparison on 2D image representation. Our SMN achieves the highest fidelity on both datasets.}
    \vspace{-0.5em}
    \label{tab:2d_results}
    \begin{tabular}{l c c c}
        \hline
        \textbf{Method} & \textbf{Kodak(dB)} & \textbf{DIV2K(dB)} & \textbf{Parameters } \\
        \hline
        MLP          & 28.63 & 30.21 & 272415 \\
        Gauss~\cite{ramasinghe2021beyond}          & 37.90 & 38.34 & 272703 \\
        SIREN ~\cite{sitzmann2020siren}         & 33.65 & 33.73 & 272703 \\
        WIRE ~\cite{saragadam2023wire}          & 40.24 & 38.90 & 265523 \\
        RINR ~\cite{ma2025robustifying}           & 32.96 & 34.03 & 289716 \\
        \textbf{SMN (Ours)} & \textbf{41.40} & \textbf{42.53} & 264216 \\
        \hline
    \end{tabular}
\end{table}

\begin{figure}[t]
    \centering
    \begin{subfigure}[b]{0.32\linewidth}
        \centering
        \includegraphics[width=\linewidth]{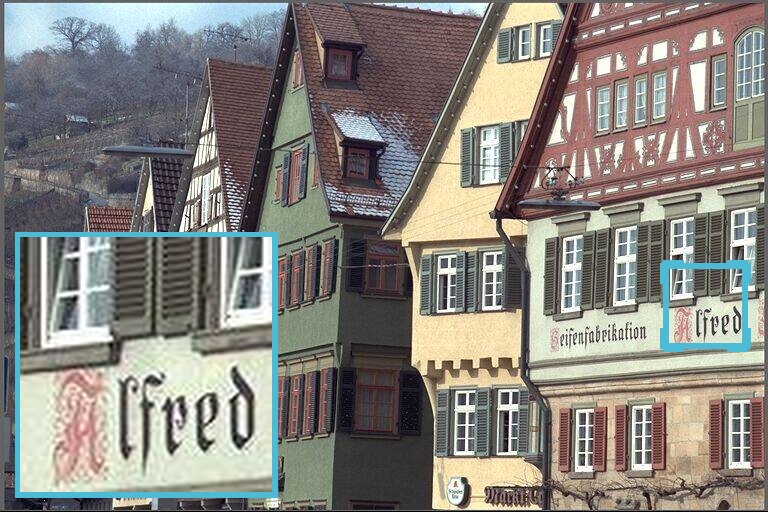}
        \caption{Ground Truth}
    \end{subfigure}\hfill
    \begin{subfigure}[b]{0.32\linewidth}
        \centering
        \includegraphics[width=\linewidth]{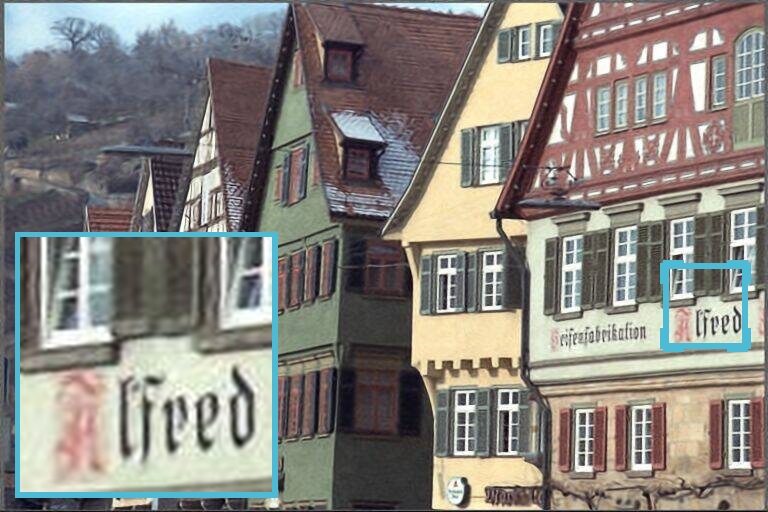}
        \caption{SIREN}
    \end{subfigure}\hfill
    \begin{subfigure}[b]{0.32\linewidth}
        \centering
        \includegraphics[width=\linewidth]{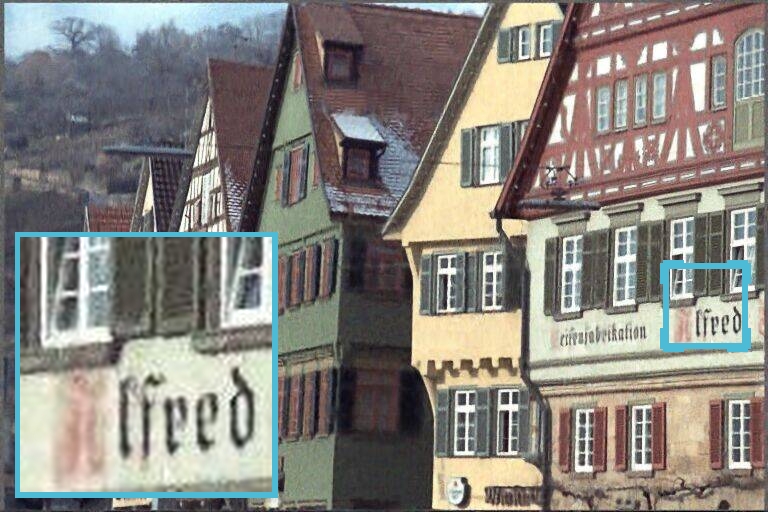}
        \caption{RINR}
    \end{subfigure}

    \medskip 

    \begin{subfigure}[b]{0.32\linewidth}
        \centering
        \includegraphics[width=\linewidth]{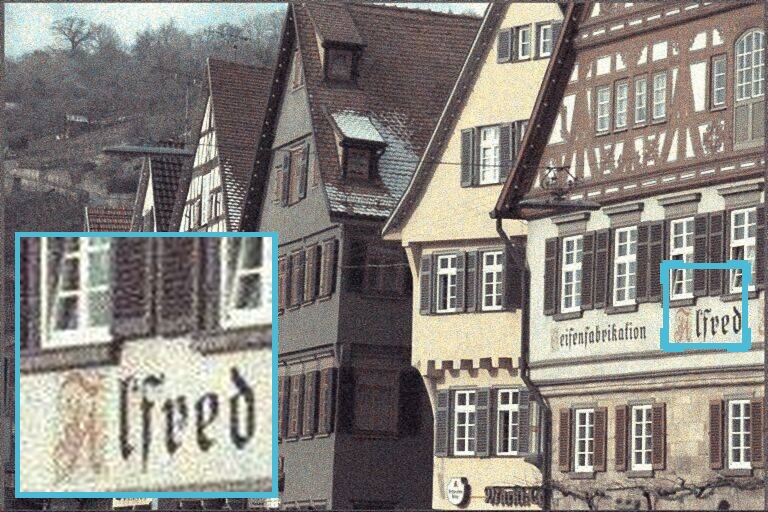}
        \caption{Gauss}
    \end{subfigure}\hfill
    \begin{subfigure}[b]{0.32\linewidth}
        \centering
        \includegraphics[width=\linewidth]{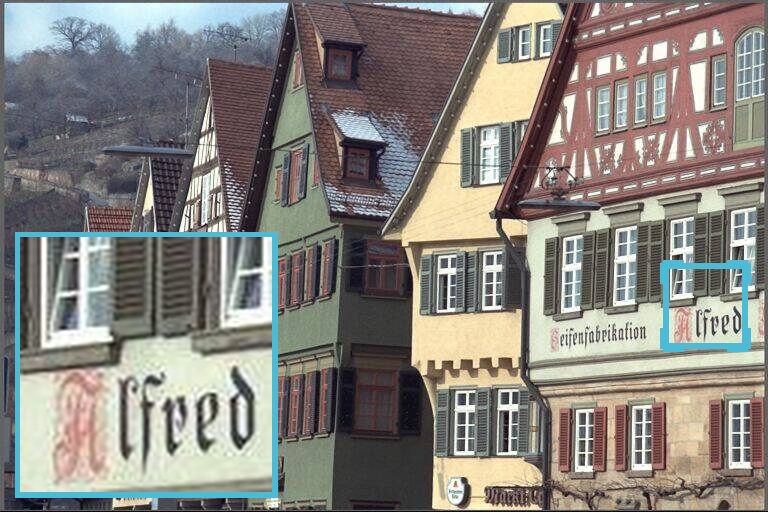}
        \caption{WIRE}
    \end{subfigure}\hfill
    \begin{subfigure}[b]{0.32\linewidth}
        \centering
        \includegraphics[width=\linewidth]{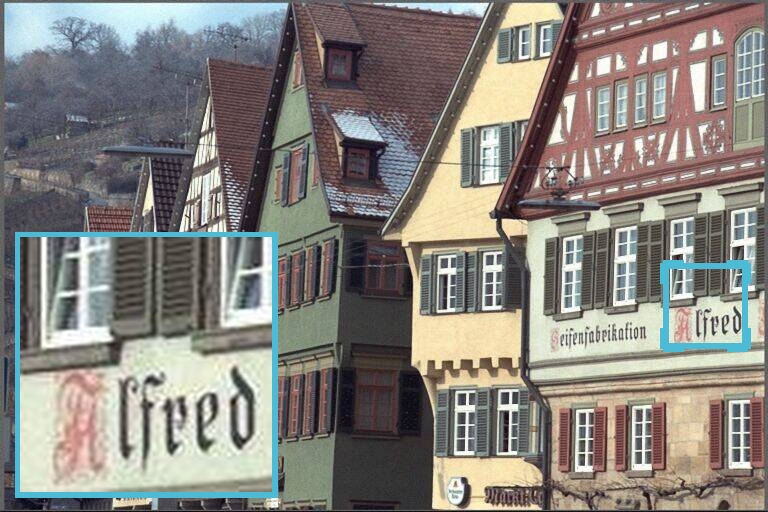}
        \caption{\textbf{SMN (Ours)}}
    \end{subfigure}

    \vspace{-1em}

    \caption{Visual comparison of reconstruction quality for different INR methods on a Kodak image. 
    Our method preserves fine textures and edges most faithfully. Best viewed on screen when zooming in.}
    \label{fig:qualitative_comparison_grid}
    \vspace{-1em}
\end{figure}

The results demonstrate the effectiveness and robustness of our SMN. On the standard Kodak benchmark, our model achieves a state-of-the-art PSNR of \textbf{41.40 dB}, surpassing the strong WIRE baseline. This strong performance is replicated on the DIV2K dataset, where the SMN again attains the highest fidelity. This is accomplished with the most compact architecture among the top-performing models. This dual advantage in both accuracy and efficiency validates that our structured, signal-processing-inspired design provides a more effective parameterization for representing natural images.This parameter efficiency directly translates to computational efficiency: inference FLOPs for a Kodak image are 208 GFLOPs for SMN, comparable to SIREN (214 G) and significantly lower than WIRE (835 G).

\subsection{Generalization to 3D Scene Representation}
\label{subsec:results_3d}

To assess our architecture's generalization, we evaluated it on the NeRF benchmark. We use the code base of ~\cite{mildenhall2021nerf}. The average PSNR across all eight scenes is reported in Table~\ref{tab:nerf_results}. This experiment provides a controlled comparison of the core network backbones, as all models utilize the same PE layer.

\begin{table}[h!]
    \centering
    \caption{Average PSNR (dB) on the eight scenes of the NeRF synthetic dataset at 400$\times$400 resolution.}
    \label{tab:nerf_results}
    \begin{tabular}{l c c}
        \hline
        \textbf{Method} & \textbf{PSNR (dB)} & \textbf{Parameters} \\
        \hline
        PE+Gauss~\cite{ramasinghe2021beyond}          & 32.00 & 287749 \\
        PE+SIREN~\cite{sitzmann2020siren}          & 29.06 & 287749 \\
        PE+WIRE~\cite{saragadam2023wire}           & 25.14 & 283479 \\
        PE+RINR~\cite{ma2025robustifying}           & 26.84 & 314703 \\
        PE+MLP~\cite{mildenhall2021nerf}            & 26.66 & 290370 \\
        \textbf{PE+SMN (Ours)} & \textbf{32.98} & 287749 \\
        \hline
    \end{tabular}
    \vspace{-1em}
\end{table}


Our \textbf{PE+SMN} model achieves an average PSNR of \textbf{32.98 dB}, significantly outperforming all baselines by a large margin (over 0.98 dB compared to the next-best). This result is critical as it isolates the contribution of our core architecture. It demonstrates that the modulative filtering mechanism is a more powerful feature processor than standard MLPs.

\begin{figure}[!t]
    \centering

    \begin{subfigure}[b]{0.32\linewidth}
        \centering
        \includegraphics[width=\linewidth]{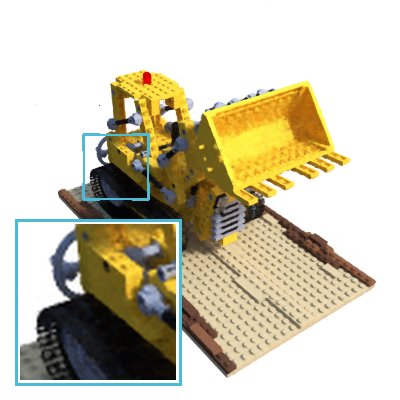}
        \caption{PE + Gauss}
    \end{subfigure}\hfill
    \begin{subfigure}[b]{0.32\linewidth}
        \centering
        \includegraphics[width=\linewidth]{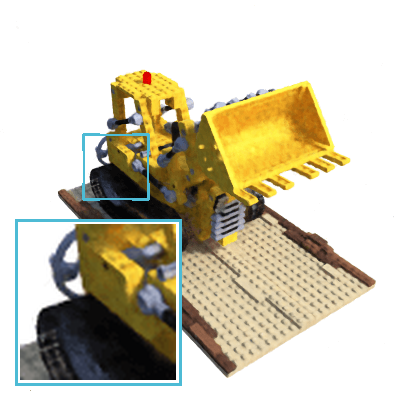}
        \caption{PE + SIREN}
    \end{subfigure}\hfill
    \begin{subfigure}[b]{0.32\linewidth}
        \centering
        \includegraphics[width=\linewidth]{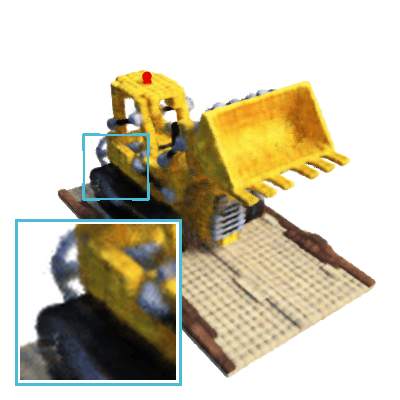}
        \caption{PE + WIRE}
    \end{subfigure}

    \medskip

    \begin{subfigure}[b]{0.32\linewidth}
        \centering
        \includegraphics[width=\linewidth]{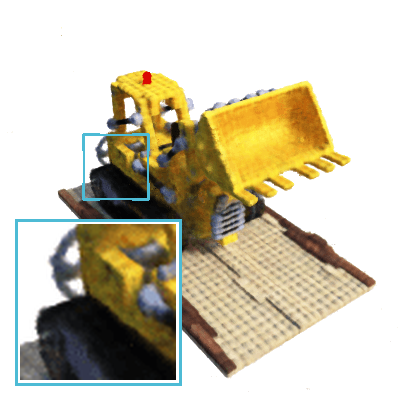}
        \caption{PE + MLP}
    \end{subfigure}\hfill
    \begin{subfigure}[b]{0.32\linewidth}
        \centering
        \includegraphics[width=\linewidth]{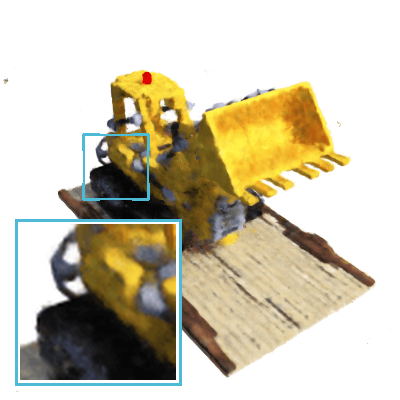}
        \caption{PE + RINR}
    \end{subfigure}\hfill
    \begin{subfigure}[b]{0.32\linewidth}
        \centering
        \includegraphics[width=\linewidth]{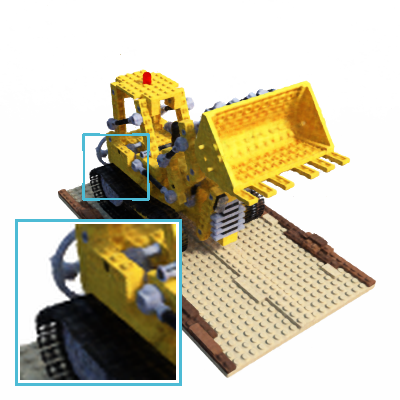}
        \caption{\textbf{PE + SMN}}
    \end{subfigure}

    \vspace{-0.5em}

    \caption{Qualitative comparison of view synthesis on the Lego NeRF dataset. 
    Our method reconstructs fine geometric details more faithfully and reduces common artifacts such as floater noise and blurriness. Best viewed on screen when zooming in.}
    \label{fig:nerf_qualitative_comparison_grid}
    \vspace{-1em}
\end{figure}

\subsection{Ablation Studies and Analysis}
\label{subsec:ablations}

To validate our key design choices and understand the sources of the SMN's performance, we conducted a series of targeted ablation studies on the Kodak dataset. We first investigated the core modulation mechanism, followed by a detailed analysis of our learnable oscillator design.

A central hypothesis of our work is that multiplicative interactions are critical for spectral sculpting in deeper network layers. We empirically validated this by creating an ablated variant, ``SMN-Add'', which replaces the core multiplicative masking operation with element-wise addition. This design choice results in a substantial performance degradation of \textbf{1.15 dB} (from 41.40 dB down to 40.25 dB). This result provides direct empirical evidence that multiplication is a key mechanism for the harmonic generation necessary to represent the fine-grained details.

We further analyzed the design of our Learnable Sine Layer to determine the optimal trade-off between complexity and expressivity. We evaluated several variants of its custom activation function, $\Phi(z)$, by varying the number of sinusoidal bases and the learnability of their amplitudes. The results, presented in 
Table~\ref{tab:ablation_oscillator}, demonstrate the superiority of our final design.


\begin{table}[h!]
    \centering
    \caption{Ablation study of the Learnable Sine Layer's activation function. Performance is measured by average PSNR (dB) on the Kodak dataset.}
    \label{tab:ablation_oscillator}
    \begin{tabular}{l l c}
        \hline
        \textbf{Variant ID} & \textbf{Description} & \textbf{PSNR (dB)} \\
        \hline
        Variant 1 & Fixed Amplitudes & 35.08 \\
        Variant 2 &  K=1 & 42.87 \\
        Variant 3 &  K=2 & 43.09 \\
        \textbf{Final Design} & \textbf{K=3} & \textbf{43.68} \\
        \hline
    \end{tabular}
\end{table}


The analysis of Table~\ref{tab:ablation_oscillator} yields several key insights. First, the  performance gap between Variant 1 and all others confirms the critical importance of learnable amplitudes; merely providing a multi-frequency basis is insufficient without the flexibility to adaptively weigh each component. Second, performance progressively increases with the number of learnable bases (from K=1 to K=3), validating our choice of a richer basis. Finally, the control experiment demonstrates that this performance gain is a direct result of our principled oscillator design, and not simply an effect of increased network depth. The results reliably point to our final design---a linear combination of three sinusoidal bases with learnable amplitudes---as the most effective and reliable choice\footnote{These  ablation results for the oscillator were obtained with a hidden dimension of 312.}.

Investigating filter depth, we found two modules optimal (41.40 dB). Three layers caused a drop to 39.63 dB due to optimization issues like gradient vanishing, while four layers only partially recovered performance (40.76 dB). Thus, the two-layer design strikes the best balance between expressivity and trainability.

\section{CONCLUSION}
\label{sec:con}

In this work, we introduced the Subtractive Modulative Network (SMN), a novel implicit neural representation framework inspired by classical signal processing. By designing a structured architecture with a learnable, multi-frequency oscillator and a series of multiplicative filter modules.
Our experiments show that the oscillator design brings significant improvement by adding only a few learnable parameters. The filters achieve additional improvement with no parameter overhead. 
The proposed SMN framework thus stands as a compelling and effective alternative to monolithic MLP designs, offering a promising path towards more efficient, interpretable, and spectrally-aware neural representations.


\bibliographystyle{IEEEbib}
\bibliography{refs}

\end{document}